\documentclass[letterpaper]{article} 
\usepackage{aaai2026}  
\usepackage{times}  
\usepackage{helvet}  
\usepackage{courier}  
\usepackage[hyphens]{url}  
\usepackage{graphicx} 
\usepackage{natbib}  
\usepackage{caption} 
\usepackage{algorithm}
\usepackage{algorithmic}

\usepackage{newfloat}
\usepackage{listings}
\DeclareCaptionStyle{ruled}{labelfont=normalfont,labelsep=colon,strut=off} 
\floatstyle{ruled}
\newfloat{listing}{tb}{lst}{}
\floatname{listing}{Listing}
\usepackage{xspace}
\usepackage{booktabs}
\usepackage{amssymb}
\usepackage{bm}
\usepackage{amsmath}
\usepackage{multirow}
\usepackage{makecell}
\usepackage{subcaption}
\usepackage[colorlinks=true, urlcolor=blue, citecolor=black]{hyperref}
\usepackage[capitalize]{cleveref}
\crefname{section}{Sec.}{Secs.}
\Crefname{section}{Section}{Sections}
\crefname{table}{Tab.}{Tabs.}
\Crefname{table}{Table}{Tables}

\newcommand{\method}{PanFlow\xspace}

\newcommand{\firstcolor}{red!30}
\newcommand{\firstcell}[1]{\cellcolor{\firstcolor}#1}
\newcommand{\secondcolor}{orange!30}
\newcommand{\secondcell}[1]{\cellcolor{\secondcolor}#1}
\newcommand{\thirdcolor}{yellow!30}
\newcommand{\thirdcell}[1]{\cellcolor{\thirdcolor}#1}
\newcommand{\graycell}[1]{\cellcolor{gray!10}#1}

\title{\method: Decoupled Motion Control for Panoramic Video Generation}
\author{
    Cheng Zhang\textsuperscript{\rm 1,2},
    Hanwen Liang\textsuperscript{\rm 3},
    Donny Y. Chen\textsuperscript{\rm 1},
    Qianyi Wu\textsuperscript{\rm 1}, \\
    Konstantinos N. Plataniotis\textsuperscript{\rm 3},
    Camilo Cruz Gambardella\textsuperscript{\rm 2,4},
    Jianfei Cai\textsuperscript{\rm 1}
}
\affiliations{

    \textsuperscript{\rm 1}Department of Data Science and Artificial Intelligence, Monash University, Clayton, Victoria, Australia
    \\
    \textsuperscript{\rm 2}Building 4.0 CRC, Caulfield East, Victoria, Australia
    \\
    \textsuperscript{\rm 3}The Edward S Rogers Sr. ECE Department, University of Toronto, Toronto, M5S3G8, Canada
    \\
    \textsuperscript{\rm 4}Future Building Initiative, Monash University, Caulfield East, Victoria, Australia
    \\
    
    \{cheng.zhang,qianyi.wu,camilo.cruzgambardella,jianfei.cai\}@monash.edu,
    donny.chen@outlook.sg,
    \\
    hw.liang@mail.utoronto.ca,
    kostas@ece.utoronto.ca
}

\usepackage{bibentry}

\begin{document}

\maketitle

\begin{abstract}

Panoramic video generation has attracted growing attention due to its applications in virtual reality and immersive media.
However, existing methods lack explicit motion control and struggle to generate scenes with large and complex motions.
We propose \textit{\method}, a novel approach that exploits the spherical nature of \underline{pan}oramas to decouple the highly dynamic camera rotation from the input optical \underline{flow} condition, enabling more precise control over large and dynamic motions.
We further introduce a spherical noise warping strategy to promote loop consistency in motion across panorama boundaries.
To support effective training, we curate a large-scale, motion-rich panoramic video dataset with frame-level pose and flow annotations.
We also showcase the effectiveness of our method in various applications, including motion transfer and video editing.
Extensive experiments demonstrate that \method significantly outperforms prior methods in motion fidelity, visual quality, and temporal coherence.
Our code, dataset, and models are available at \url{https://github.com/chengzhag/PanFlow}.
\end{abstract}


\section{Introduction}\label{sec:intro}

Generating realistic and immersive videos depicting the world around us is among the most compelling challenges in content creation.
Panoramic videos, in particular, provide a natural representation by offering continuous 360° views that fully capture the surrounding environment, making them ideal for applications in virtual reality, immersive storytelling, and cinematic content creation.
Enabled by advancements in diffusion-based generative models and the availability of panoramic datasets~\cite{wang2024360dvd}, recent works have shown promising results in synthesizing photorealistic panoramic scenes from simple text or image prompts~\cite{liu2025dynamicscaler,zhou2025holotime}.

\begin{figure}[t]
    \centering
    \includegraphics[width=0.98\linewidth, trim={0px 0px 0px 0px}, clip]{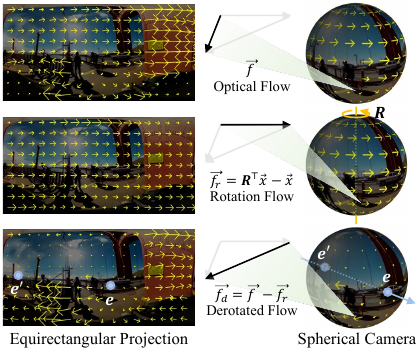}
    \caption{
        \textbf{Spherical Camera Optical Flow.}
        The optical flow from a panoramic video (left) can be interpreted as a spherical camera optical flow (right).
        For complex motion $\vec{f}$, the camera rotation yields an analytic rotation flow $\vec{f}_r$ on the sphere.
        By decomposing $\vec{f}$ into $\vec{f}_r$ and its residual, we obtain a derotated flow $\vec{f}_d$ that more clearly captures camera translation and object dynamics.
    }\label{fig:flow}
\end{figure}

Despite these advances, precise motion control in panoramic video generation remains an open challenge~\cite{wang2024360dvd,wu2024draganything}.
Unlike traditional perspective videos, panoramic videos require loop consistency across boundaries, both in appearance and motion.
Motion in panoramic videos involves complex interactions between two distinct yet intertwined components: \textit{object motion} and \textit{camera motion}.
Accurate and coherent control of these motion components is crucial for an immersive viewing experience.
Any inaccuracies can introduce noticeable artifacts, such as unnatural object trajectories, discontinuous motion across panorama seams, severely degrading the realism and practicality of generated content.

In this paper, we address this critical challenge by proposing \method, a novel motion-controllable panoramic video generation framework conditioned on optical flow.
We exploit the spherical nature of panoramic imagery and analytically decompose the input optical flow into two fundamental components: a deterministic \textit{rotation flow} induced by camera rotation on the sphere, and a residual \textit{derotated flow} that captures camera translations and object dynamics (\cref{fig:flow}).
This decomposition is not only theoretically grounded, as camera-induced rotation flow is independent of scene content in panoramic view~\cite{gluckman1998ego}, but also practically beneficial.
It allows us to simplify the motion modeling by conditioning the diffusion model solely on the derotated flow, followed by an inverse derotation to recover the full motion.
This decoupled motion control mechanism enhances synthesis stability, improves motion fidelity, and facilitates more coherent and realistic generation of dynamic panoramic scenes.
To effectively condition the model on flow, we further introduce a \textit{spherical noise warping} strategy.
Instead of conditioning the model on optical flow embeddings, we warp the initial Gaussian noise across frames using optical flow fields.
Importantly, in our design, noise pixels can be propagated across panorama boundaries to ensure seamless motion continuity under spherical geometry.
Finally, to support high-quality training, we meticulously curate a large-scale dataset featuring dynamic content, diverse motion, and clean equirectangular projections, with frame-level pose and flow annotations.
Leveraging proposed techniques and the curated dataset, \method achieves state-of-the-art performance in panoramic video generation, exhibiting significantly improved motion fidelity, perceptual realism, and temporal coherence over existing approaches.
Our contributions are summarized as follows:
\begin{itemize}
\item We propose \textit{\method}, the first motion-controllable panoramic video generation framework capable of synthesizing temporally coherent 360° videos from a single image, precisely following user-defined motion flow.
\item We introduce a decoupled motion control mechanism that isolates rotation flow from translation and object motion, and develop a spherical noise warping strategy to enable flow-conditioned generation that preserves loop consistency across panorama boundaries.
\item We curate a large-scale, motion-rich panoramic video dataset with frame-level flow and pose annotations, to support training and evaluation in motion-controlled panoramic video synthesis.
\item Extensive experiments show that \method significantly outperforms prior works in motion fidelity, magnitude, and overall visual quality across multiple benchmarks.
\end{itemize}

\begin{figure*}[t]
    \centering
    \includegraphics[width=1.\linewidth, trim={0 0 0 0}, clip]{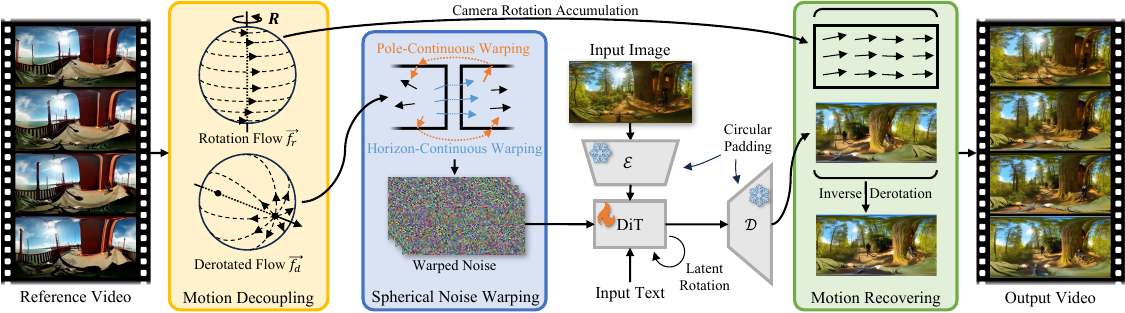}
    \caption{
        \textbf{Our proposed \method pipeline.}
        Given an input image and text prompt, \method uses a decoupled motion from a video as reference to generate a panoramic video.
        We first estimate a decoupled optical flow from the reference video, of which the derotated flow is used to generate a latent noise with spherical noise warping.
        The latent noise then serves as a motion condition for a video diffusion transformer with LoRA fine-tuning to generate derotated videos.
        Finally, the decoupled rotation is accumulated and applied to the generated video frames to recover the full motion.
    }\label{fig:pipeline}
\end{figure*}

\section{Related Work}\label{sec:related_work}

\subsection{Video Diffusion Models}
Diffusion-based generative models have significantly improved the realism and diversity of synthesized videos.
Early efforts extended image diffusion models~\cite{rombach2022high, guo2024animatediff} with temporal modules, while recent approaches~\cite{yang2025cogvideox, wan2025wan} embed videos with 3D-VAE and leverage diffusion transformers for denoising.
For controllability, most models incorporate conditioning signals such as texts or reference images~\cite{ho2022video, yang2025cogvideox}.
Beyond semantic controls, some works employ structural guidance like depth maps or edge cues~\cite{wang2023videocomposer, guo2024sparsectrl}.
Our method focus on motion control by leveraging synthetic optical flow as an explicit, scene-agnostic motion condition, enabling precise and temporally coherent control over dynamic content.

\subsection{Panoramic Image and Video Generation}
More efforts have adapted diffusion models to panoramic domains for immersive scene synthesis.
PanoGen~\cite{li2023panogen} employed latent diffusion to synthesize indoor panoramas from text, while StitchDiffusion~\cite{wang2024customizing} introduced structural priors to enforce boundary alignment.
PanFusion~\cite{zhang2024taming} proposed a dual-branch design to handle global context and local detail.
In video domains, DynamicScaler~\cite{liu2025dynamicscaler} designed training-free diffusion for panoramic and arbitrary aspect-ratio video generation, while HoloTime~\cite{zhou2025holotime} employed 4DGS~\cite{kerbl20233d} to generate explorable 4D scenes.
PanoDiT~\cite{zhang2025panodit} and PanoWan~\cite{xia2025panowan} finetuned diffusion transformers for panoramic video generation.
However, these works lack explicit motion control and often entangle camera and object motions with noticeable artifacts.
The closest work to ours is 360DVD~\cite{wang2024360dvd}, which extended perspective video diffusion models using 360-Adapter, allowing optional motion conditions.
However, it exhibits limited motion control compared to our method, which enables fine-grained, user-customizable motion generation with superior visual quality and motion fidelity.

\subsection{Motion-conditioned Video Generation}
Precise motion control is essential for interactive and editable video synthesis. 
Prior works explored object-level control, scene-level control, and motion transfer.
Object-level control focuses on directing specific elements using inputs like masks or trajectories.
SG-I2V~\cite{namekata2025sg} and DragAnything~\cite{wu2024draganything} manipulate localized motion using learned trajectories or entity representations.
Scene-level control involves guiding global dynamics via camera trajectories, either through explicit pose supervision~\cite{he2024cameractrl,wang2024motionctrl} or coarse directional categories, e.g., panning or zooming~\cite{guo2024animatediff}.
Motion transfer aims to adapt temporal dynamics from reference videos to new content.
MotionClone~\cite{ling2025motionclone} introduced attention-based motion encoding to improve transferability.
Go-with-the-Flow~\cite{burgert2025go} leveraged warped noise for detailed motion control.
While effective, these methods are limited to perspective videos and falls short to generalize to the spherical topology of panoramic content, which requires structural consistency across seams and poles.
We propose the first framework for motion-controlled panorama-to-video generation, introducing tailored spherical noise warping and decoupled motion, enabling loop consistency and better motion fidelity.

\section{Method}\label{sec:method}

\method is a motion-controlled video generation framework that synthesizes dynamic panoramic scenes from an input panorama.
The core of \method is a fine-tuned video diffusion model that takes in spherical warped noise encoded for motion conditions.
To enhance fidelity in motion control, we introduce a motion decoupling mechanism to separate camera rotation from other motion components. 
Additionally, we construct a specialized dataset for training \method, composed of highly dynamic panoramic videos and diverse motion patterns.
The overall framework is shown in \cref{fig:pipeline}.

\subsection{Loop Consistent Motion-Controllable Panoramic Video Diffusion}
Unlike conventional video diffusion models that generate videos in a perspective view, \method targets the generation of 360° panoramic videos.
Direct application of perspective video diffusion techniques to panoramas often results in loop inconsistencies, particularly related to motion, semantics, and low-level pixels.
Specifically, moving objects exiting boundaries of the panorama must seamlessly re-enter from the opposite sides (\textit{motion consistency}), latent representations must maintain semantic coherence across panorama boundaries (\textit{semantic consistency}), and decoded videos in pixel-space must form a seamless loop at the boundaries (\textit{pixel consistency}).
To ensure these consistencies, we introduce separate techniques to address each aspect.

\subsubsection{Spherical Noise Warping.}
To enable effective motion control in panoramic video generation, we propose a spherical noise warping technique.
Our method leverages the fact that Gaussian noise inherently defines image contents in diffusion models~\cite{burgert2025go}.
We introduce the spherical noise warping where the initial Gaussian noise is warped iteratively to subsequent timestamps using optical flow from a source video.
This warped noise maintains distributional similarity across corresponding regions in different frames, achieving both temporal coherence and motion control.
Formally, given Gaussian noise in a latent space of size $H \times W$, we denote previous-frame noise as $\bm{q}\in\mathbb{R}^{D}$ (where $D=H\times W$) and use noise warping operator~\cite{burgert2025go} to generate next-frame noise~$\bm{q}'$ with forward and backward optical flows $\bm{f}, \bm{f}^{\prime}: D \rightarrow \mathbb{N}^2$.
To ensure seamless loop motion consistency specifically for panoramic content, noise warping crossing panorama boundaries is performed by defining target pixels $(\hat{i}, \hat{j})$ as:
\begin{equation}
\begin{aligned}
\hat{i} &=
\begin{cases}
-\,i, & i < 0,\\
i, & 0 \le i < H,\\
2(H-1)-i, & i \ge H,
\end{cases}\\[1ex]
\hat{j} &=
\begin{cases}
\bigl(j + \tfrac{W}{2}\bigr)\bmod W, & i<0 \text{ or } i\ge H,\\
j \bmod W, & 0 \le i < H,
\end{cases}
\end{aligned}
\end{equation}
where $(i, j)$ are the pixel coordinates mapped by the optical flow.
This operation passes the content across the left-right boundaries and poles based on the connectivity of the equirectangular projection.
We employ this specially warped noise to fine-tune a diffusion transformer equipped with a LoRA adapter~\cite{hu2022lora}, as illustrated in \cref{fig:pipeline}.

\subsubsection{Latent Rotation on 3D Latent Maps.}
We build upon the latest video DiT model~\cite{yang2025cogvideox} that leverages transformers for denoising.
Previous works~\cite{zhang2024taming,wu2023ipo} have shown that the pre-trained priors in the diffusion model conflict with the periodic structure of the panorama, tending to break the loop consistency in the latent space.
To mitigate this problem, we apply latent rotation before tokenization on the 3D latent map.
Specifically, at each denoising step, we rotate the latent map along the longitude axis
with a fixed angle of ${\theta}$ by rolling for ${\theta} / 360 \times {W}$ latent pixels along the ${W}$ dimension.
The rolling distance is accumulated during the denoising process to rotate the denoised result back to the original orientation.
This rotation helps DiT to blend the seam after it is shifted into the frame, instead of concentrating discontinuities at the boundaries.

\subsubsection{Circular Padding.}
Inspired by~\cite{zhuang2022acdnet}, we integrate circular padding into both encoder $\mathcal{E}$ and decoder $\mathcal{D}$ of the 3D-VAE to maintain pixel-level loop consistency in both latent maps and decoded videos.
Specifically, we pad the left-right boundaries with latents or pixels from opposite sides, enabling convolution kernels to operate seamlessly across panorama edges.
After encoding or decoding, outputs are cropped back to their original dimensions.
The padding width is set to 8 pixels for the encoder and 1 latent pixel for the decoder, sufficient to cover the receptive field.

\subsection{Decoupled Motion Control}

Although the proposed techniques above establish a strong baseline for motion control in panoramic video generation, they still face challenges in highly dynamic scenarios.
We observe that the motion condition often comprises a complex combination of camera rotation, translation, and object motions.
Considering the spherical nature inherent to panoramic videos, we introduce a decoupled motion control to isolate camera rotation from other motion components.
This decoupling reduces the modelling burden over the video diffusion model, thereby enhancing its ability to effectively follow complex motion conditions.

\subsubsection{Spherical Camera Optical Flow.}
A 360° panorama corresponds to an equirectangular projection (\cref{fig:flow} left) from a spherical camera (\cref{fig:flow} right).
Camera motions can thus be represented using spherical optical flow~\cite{gluckman1998ego}.
Specifically, for a motion described by optical flow $\vec{f}$, one independent component is its camera rotation.
When the camera rotation is defined as rotation matrix $\bm{R} \in \mathbb{R}^{3 \times 3}$, its optical flow vector $\vec{f} \in \mathbb{R}^3$ at pixel position $\vec{x} = ({x}, {y}, {z}) \in \mathbb{R}^3$ on the unit sphere can be calculated as:
\begin{equation}
    \vec{f}_{r} = \bm{R}^\top \vec{x} - \vec{x}.
\end{equation}
This rotation flow follows a circular pattern around the rotation axis, and the magnitude is proportional to the distance to the rotation axis, as shown in the second row of \cref{fig:flow}.
In contrast, in a static scene, when the camera moves with only translation, the optical flow vectors diverge from one epipole $\bm{e}$ and converge to opposite epipole $\bm{e}^\prime$, where $\bm{e} - \bm{e}^\prime$ is the direction of the camera translation and the magnitude of the flow vector depends on the scene depth.

\subsubsection{Decoupling Camera Rotation from Motion Condition.}
We note that the rotation flow is invariant to the scene content, which can be easily decoupled from the input optical flow with a derotation operation:
\begin{equation}
    \vec{f}_{d} = \vec{f} - \vec{f}_{r}.
\end{equation}
The decoupling operation is simple yet effective, as it allows the video diffusion model to focus on the derotated flow, which only contains translation and object motion, while the camera rotation is handled separately.
In practice, we directly derotate the input video and then estimate the derotated optical flow with a panoramic flow estimator~\cite{shi2023panoflow} for more stable optical flow conditions, when input video is available in most applications such as motion transfer and object editing.

\subsubsection{Recovering Camera Rotation for Generated Videos.}
After the derotated video is generated with the derotated optical flow as condition, we recover the decoupled camera rotation by applying the inverse derotation operation on each frame $\bm{I}_{t}$:
\begin{equation}
    \hat{\bm{I}}_{t} = \text{RotateERP}(\bm{I}_{t}, \bm{R}_{t})
\end{equation}
where $\bm{R}_{t}$ is the rotation matrix relative to the first frame, and $\text{RotateERP}(\cdot)$ is the operation that rotates ERP panoramas with bilinear interpolation so that $\hat{\bm{I}}_{t}(\vec{x}) = \text{Bilinear}(\bm{I}_{t}, \bm{R}_{t}^\top \vec{x})$ for each pixel $\vec{x}$ on the unit sphere of $\bm{I}_{t}$.
$\bm{R}_{t}$ can be extracted from the input video with SLAM~\cite{sumikura2019openvslam} or from the input optical flow by accumulating estimated rotation differences~\cite{kim2021self}.
We note that this operation is achievable thanks to the omnidirectional nature of the panorama, which allows us to rotate the panorama arbitrarily with minimal information loss.

\begin{figure*}[t]
    \centering
    \includegraphics[width=1.\linewidth, trim={0 0 0 0}, clip]{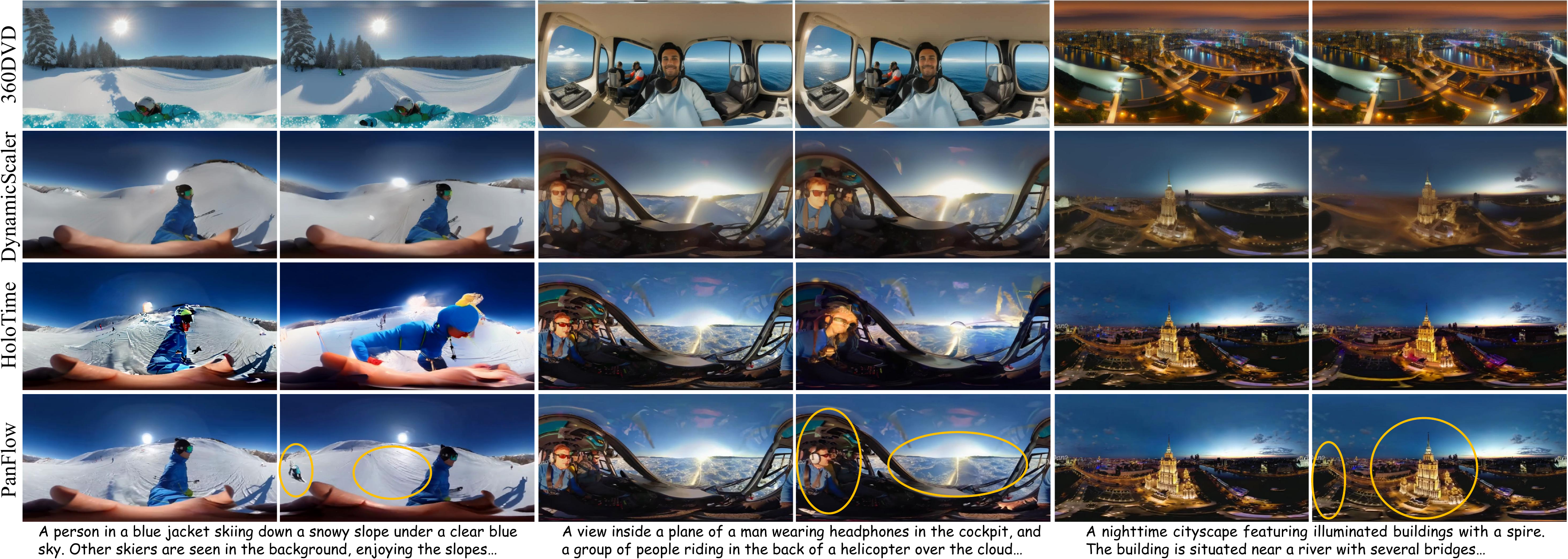}
    \caption{
        \textbf{Comparison with Panoramic Video Generation Methods.} 
        We compare our proposed \method with the baselines conditioned on the same input images and text prompts (360DVD uses text prompts only).
        Generated frames are shown at the same timestamps.
        We highlight regions exhibiting more dynamic motion, high-fidelity textures, and consistent geometry.
    }\label{fig:pano}
\end{figure*}

\subsection{Motion-Rich Dataset Construction}
Training an effective motion-controllable panoramic diffusion model requires a highly dynamic motion-rich training dataset, including diverse camera motions and object motions.
However, existing datasets such as WEB360~\cite{wang2024360dvd} exhibit limited motion diversity and magnitude, often featuring static cameras and minimal object movement.
To address this, we construct a new dataset by curating a collection of videos from a large corpus dataset, 360-1M~\cite{wallingford2024image}.
We design a specialized preprocessing pipeline tailored for panoramic videos to filter, segment, and annotate clips for training \method, including:
\begin{itemize}
    \item \textbf{Format Check.} 360-1M contains a portion of videos that are not in the equirectangular projection format.
    We filter out these videos with handcrafted heuristics, such as computing SSIM~\cite{ssim} between the left and right halves to exclude stereo (3D) formats, and applying spherical masks to identify and remove fisheye footage.
    \item \textbf{Transition Detection.} We apply the \texttt{scenedetect} \cite{scenedetect2024} library to identify scene transitions, including hard cuts and black fades.
    \item \textbf{SLAM for Clip and Pose Extraction.} We use a panoramic visual SLAM~\cite{sumikura2019openvslam} to extract the camera poses. As some fade cuts can not be easily detected by the transition detection, we exploit the ``tracking lost'' signal from the SLAM to further segment the scene clips.
    \item \textbf{Watermark Filtering.} We apply a deep learning model~\cite{watermark} to detect watermarks and captions in the videos, and filter out videos with high watermark scores.
    \item \textbf{Motion Filtering.} To enhance the dynamics of dataset, we compute the optical flow~\cite{opencv_library}, and discard clips with low flow magnitudes and filter out those with minimal camera motion based on estimated poses.
\end{itemize}
The resulting dataset offers a rich and diverse set of 150k, 3-10s dynamic panoramic clips, with 53\% natural, 22\% urban, 9\% indoor, and 16\% others.
We detail the dataset construction process in the supplementary material.

\section{Experiments}\label{sec:exp}

\begin{table*}[t]
    \centering
    \newcommand{\dataset}[1]{\multirow{6}{*}{\rotatebox[origin=c]{90}{#1}}}
    \resizebox*{1.0\linewidth}{!}{
        \begin{tabular}{rlcccccccccc}
        \toprule
         & \multirow[c]{2}{*}{Method} & \multirow[c]{2}{*}{FVD$\downarrow$} & \multirow[c]{2}{*}{FID$\downarrow$} & \multirow[c]{2}{*}{CLIP$\uparrow$} & \multicolumn{4}{c}{Motion Control} & \multicolumn{3}{c}{Q-Align} \\
        \cmidrule(lr){6-9} \cmidrule(lr){10-12}
         &  &  &  &  & Flow EPE$\downarrow$ & LPIPS$\downarrow$ & PSNR$\uparrow$ & SSIM$\uparrow$ & Image Quality$\uparrow$ & Image Aesthetic$\uparrow$ & Video Quality$\uparrow$ \\
        \midrule
        \dataset{360-1M} & 360DVD & 1663.61 & 152.04 & 24.57 & \thirdcell 4.467 & 0.771 & 9.41 & 0.319 & 0.5426 & \thirdcell 0.4231 & \thirdcell 0.6073 \\
        & DynamicScaler & 1360.94 & 85.25 & 25.14 & 8.647 & 0.529 & 14.32 & \thirdcell 0.500 & 0.5137 & 0.4142 & 0.5884 \\
        & HoloTime & \thirdcell 1013.32 & \thirdcell 59.59 & 24.62 & 6.322 & \thirdcell 0.443 & \thirdcell 14.83 & 0.490 & 0.5218 & 0.4016 & 0.5929 \\
        & MotionClone & 1435.66 & 87.60 & \thirdcell 25.71 & 9.613 & 0.577 & 11.91 & 0.412 & \secondcell 0.5533 & \secondcell 0.4364 & \secondcell 0.6147 \\
        & GoWithTheFlow & \secondcell 477.13 & \secondcell 28.59 & \secondcell 25.75 & \secondcell 3.297 & \secondcell 0.306 & \secondcell 18.43 & \secondcell 0.622 & \thirdcell 0.5459 & 0.4069 & 0.5879 \\
        & PanFlow & \firstcell 298.39 & \firstcell 23.01 & \firstcell 27.14 & \firstcell 2.011 & \firstcell 0.243 & \firstcell 21.21 & \firstcell 0.723 & \firstcell 0.5552 & \firstcell 0.4397 & \firstcell 0.6196 \\
        \midrule
        \dataset{WEB360} & 360DVD & 1225.55 & 153.55 & \thirdcell 25.60 & 1.116 & 0.726 & 11.10 & 0.416 & \secondcell 0.6918 & \secondcell 0.5980 & 0.7342 \\
        & DynamicScaler & 979.30 & 94.84 & 25.30 & 3.132 & 0.464 & 18.20 & 0.634 & 0.5991 & 0.5478 & 0.6880 \\
        & HoloTime & \thirdcell 443.56 & \thirdcell 40.85 & 25.31 & \thirdcell 0.899 & \thirdcell 0.250 & \thirdcell 23.97 & \thirdcell 0.771 & \thirdcell 0.6865 & 0.5714 & \thirdcell 0.7407 \\
        & MotionClone & 667.19 & 72.53 & \secondcell 25.71 & 1.864 & 0.405 & 17.03 & 0.634 & 0.6831 & \thirdcell 0.5784 & \secondcell 0.7466 \\
        & GoWithTheFlow & \secondcell 319.01 & \secondcell 23.81 & 25.30 & \secondcell 0.710 & \secondcell 0.219 & \secondcell 25.38 & \secondcell 0.795 & 0.6357 & 0.5187 & 0.6743 \\
        & PanFlow & \firstcell 195.77 & \firstcell 19.44 & \firstcell 26.84 & \firstcell 0.486 & \firstcell 0.185 & \firstcell 27.46 & \firstcell 0.842 & \firstcell 0.7014 & \firstcell 0.6165 & \firstcell 0.7515 \\
        \bottomrule
        \end{tabular}
    }
    \caption{
        \textbf{Comparison with SoTA Methods.}
        We evaluate the video generation quality with FVD, FID, CLIP scores, and user-centric Q-Align metrics.
        For motion control, we report Flow EPE (End-Point Error), LPIPS, PSNR, and SSIM scores.
        \method achieves consistently superior performance across both 360-1M and WEB360 datasets.
    }\label{tbl:cmp}
\end{table*}

\begin{figure*}[t]
    \centering
    \includegraphics[width=1.\linewidth, trim={0 0 0 0}, clip]{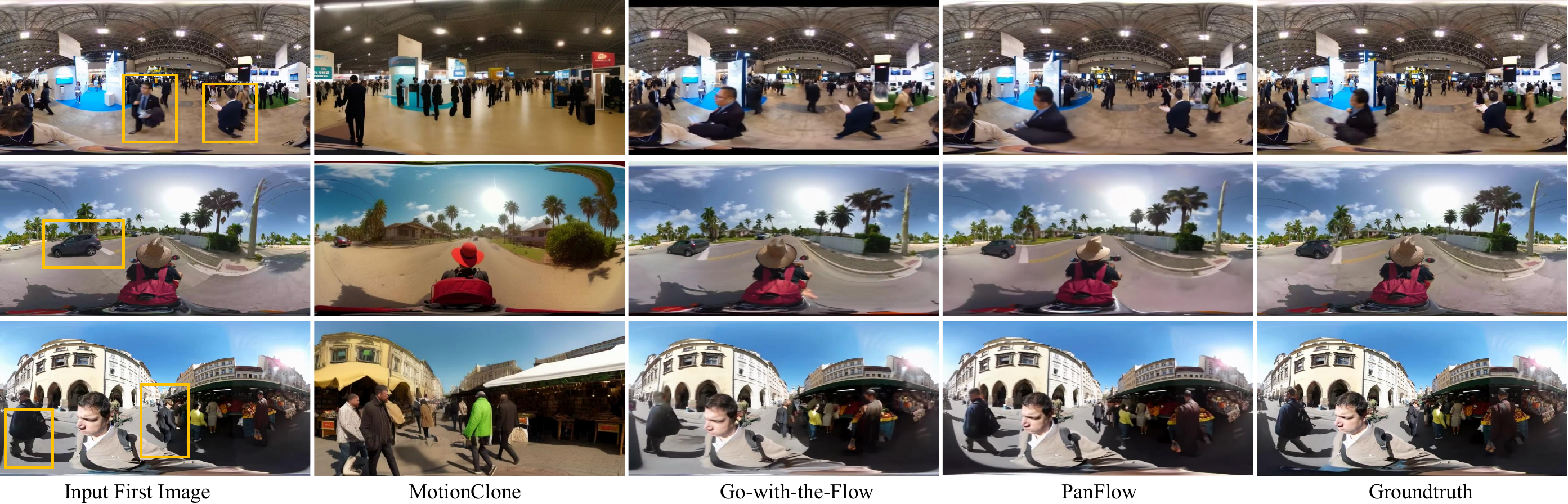}
    \caption{
        \textbf{Comparison with Motion-controlled Video Generation Methods.}
        All the methods generate videos conditioned on the same input images and motion flows.
        \method better follows the flow conditions and aligns more closely with the ground truth.
        We use rectangles to highlight regions with large motion.
         (Text prompts omitted without loss of generality.)
    }\label{fig:motion}
\end{figure*}

\begin{table*}[t]
    \centering
    \resizebox*{1.0\linewidth}{!}{
        \begin{tabular}{lclcccccccc}
        \toprule
        \multirow[c]{2}{*}{Method} & \multirow[c]{2}{*}{\makecell[c]{Decoupled \\ Motion}} & \multirow{2}{*}{Loop Consistency Setup} & \multirow[c]{2}{*}{\makecell[c]{End \\ Continuity$\downarrow$}} & \multirow[c]{2}{*}{FVD$\downarrow$} & \multirow[c]{2}{*}{FID$\downarrow$} & \multirow[c]{2}{*}{CLIP$\uparrow$} & \multicolumn{4}{c}{Motion Control} \\
        \cmidrule(lr){8-11}
        &  &  &  &  &  &  & Flow EPE$\downarrow$ & LPIPS$\downarrow$ & PSNR$\uparrow$ & SSIM$\uparrow$ \\
        \midrule

        \multirow[c]{2}{*}{GWTF} & w/o & w/o & \secondcell 0.1389 & \secondcell 610.69 & \secondcell 32.34 & \firstcell 25.77 & \secondcell 5.754 & \secondcell 0.435 & \secondcell 14.63 & \secondcell 0.500 \\
        & \graycell w/ & \graycell w/o & \firstcell 0.0276 & \firstcell 580.79 & \firstcell 32.26 & \firstcell 25.77 & \firstcell 4.818 & \firstcell 0.420 & \firstcell 15.01 & \firstcell 0.520 \\

        \midrule

        \multirow[c]{5}{*}{\method} & \multirow{4}{*}{w/o} & w/o Circular Padding & 0.0512 & \thirdcell 429.29 & \thirdcell 26.01 & \thirdcell 27.07 & \secondcell 4.254 & \thirdcell 0.370 & \thirdcell 16.86 & \secondcell 0.573 \\
        & & w/o Latent Rotation & 0.0287 & \secondcell 428.58 & 26.31 & \secondcell 27.08 & 4.306 & 0.371 & 16.76 & 0.568 \\
        & & w/o Spherical Noise Warping & \thirdcell 0.0286 & 441.59 & 26.04 & \secondcell 27.08 & 4.289 & 0.371 & 16.83 & \thirdcell 0.572 \\
        & & Full & \secondcell 0.0264 & 430.90 & \secondcell 25.89 & \thirdcell 27.07 & \thirdcell 4.260 & \secondcell 0.369 & \secondcell 16.87 & \secondcell 0.573 \\
        & \graycell w/ & \graycell Full & \firstcell 0.0257 & \firstcell 425.78 & \firstcell 25.21 & \firstcell 27.21 & \firstcell 3.626 & \firstcell 0.357 & \firstcell 17.22 & \firstcell 0.593 \\

        \bottomrule
        \end{tabular}
    }
    \caption{
        \textbf{Ablation Study.}
        We compare ablated variants of \method on 360-1M dataset.
        We also show that our proposed decoupled motion control module can be integrated into GoWithTheFlow (GWTF)~\cite{burgert2025go} as a plug-and-play module.
    }\label{tbl:ablation}
\end{table*}

\begin{figure}[t]
    \centering
    \includegraphics[width=1.\linewidth, trim={0px 0px 0px 0px}, clip]{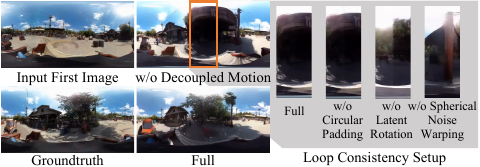}
    \caption{
        \textbf{Ablation Study.}
        Panoramas are horizontally rotated by 180 degrees to better visualize the seam.
        We zoom in on the seam to compare different loop consistency setups.
    }\label{fig:ablation}
\end{figure}

\begin{figure}[t]
    \centering
    \begin{subfigure}[t]{1.\linewidth}
        \centering
        \includegraphics[width=1.\linewidth, trim={0px 0px 0px 0px}, clip]{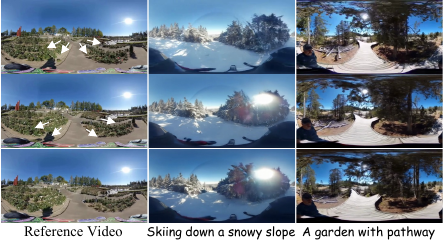}
        \caption{
            \textbf{Motion Transfer.}
        }\label{fig:transfer}
    \end{subfigure}

    \begin{subfigure}[t]{1.\linewidth}
        \centering
        \includegraphics[width=1.\linewidth, trim={0px 0px 0px 0px}, clip]{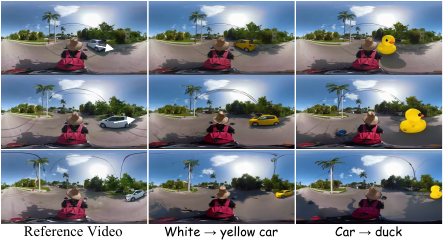}
        \caption{
            \textbf{Video Editing.}
        }\label{fig:editing}
    \end{subfigure}
    \caption{
        \textbf{Applications.}
        Our method can be applied to motion transfer and video editing tasks.
        We visualize the direction of motion in the reference video (left) with white arrows.
        Full prompts and changes are omitted for brevity.
    }\label{fig:applications}
\end{figure}

\subsection{Experimental Setup}
\subsubsection{Implementation Details.}
We hold out 100 clips from our curated dataset as a test set.
To further evaluate the out-of-domain generalizability, we additionally select 100 random clips from the WEB360 dataset~\cite{wang2024360dvd} for evaluation.
Please refer to the supplementary material for more details about the training and inference details.

\subsubsection{Baselines.}
To comprehensively evaluate the effectiveness of our method, we compare against a diverse set of strong baselines covering both panoramic and flow-conditioned video generation.
For panoramic video generation, we include text-driven model 360DVD~\cite{wang2024360dvd} and image-conditioned frameworks DynamicScaler~\cite{liu2025dynamicscaler} and HoloTime~\cite{zhou2025holotime}.
To assess motion control quality, we further compare against two recent flow-conditioned works, MotionClone~\cite{ling2025motionclone} and GoWithTheFlow~\cite{burgert2025go}.
These baselines provide a broad and representative benchmark spanning different conditioning modalities and generation paradigms.

\subsubsection{Evaluation Metrics.}
Evaluation is conducted using a comprehensive suite of metrics.
Visual quality and temporal coherence of the generated videos are evaluated using Fréchet Inception Distance (FID)~\cite{heusel2017gans} and Fréchet Video Distance (FVD)~\cite{fvd}.
Text alignment is measured by CLIP~\cite{radford2021learning} similarity.
Motion controllability and appearance quality is assessed by frame-wise visual similarity measuring PSNR, SSIM~\cite{ssim}, and LPIPS~\cite{lpips} between the generated images and ground-truth views.
To specifically assess motion control fidelity, we additionally report Flow EPE~\cite{otte1994optical}, which measure the deviation between optical flows estimated from generated and ground-truth videos, isolating motion fidelity from visual quality.
For user-centric evaluation, we follow~\cite{liu2025dynamicscaler} and incorporate Q-Align~\cite{wu2024q}, an LLM-based visual evaluator that scores content based on aesthetic appeal and perceptual quality. This metric shows strong alignment with human judgment in visual assessment tasks.
When evaluating different loop consistency setups, we also report the end continuity error from~\cite{xia2025panowan} that measures MSE between the leftmost and rightmost columns of the generated panoramic video.

\subsection{Comparison with Previous Methods}

\subsubsection{Quantitative Comparison.}
\cref{tbl:cmp} presents the quantitative comparison between our method and the baselines on the 360-1M and WEB360 datasets.
\method consistently achieves the best performance across all evaluation metrics on both datasets.
In terms of generation quality, \method shows significant improvements in FVD, FID, and CLIP scores, indicating its superior video quality, temporal coherence, and text alignment.
On human-aligned Q-Align metrics, \method also outperforms all baselines, demonstrating superior aesthetics and visual quality that align closely with human preferences.
In terms of motion control, \method achieves the best performance in Flow EPE, LPIPS, PSNR, and SSIM, demonstrating its effectiveness in generating temporally coherent videos with accurate motion control.
We note that our method is, to the best of our knowledge, the first to achieve high-quality panoramic video generation from image and flow motion conditions.
Therefore, we focus our comparisons on motion-controllable video generation baselines, i.e. MotionClone and GoWithTheFlow, on these metrics.
In addition, as 360DVD generates panoramic video from motion without image condition, we compare it using Flow EPE that solely evaluates motion fidelity, showing that 360DVD falls short significantly.
In both tasks, we evaluate directly generalization to WEB360 dataset, where \method still achieves the best performance, highlighting its strong generalization capability to out-of-domain data.

\subsubsection{Qualitative Comparison.}
\cref{fig:pano} and \cref{fig:motion} present qualitative comparisons of our method with panoramic video generation and motion-controllable video generation baselines, respectively.
In \cref{fig:pano}, we show the first and last frames of the generated videos.
\method produces high-quality panoramic videos with shaper details and more dynamic motion, showing more preserved textures on the snowy slope, more realistic motion of pilot turning head, and more consistent 3D geometry of the nighttime aerial view.
In \cref{fig:motion}, we compare \method with two flow-conditioned image-to-video generation methods, MotionClone and GoWithTheFlow.
The image condition is shown on the left, and the generated 16th frames are shown on the right.
\method demonstrates more accurate motion control, with generated frames more closely matching the target frame.
It also produces more coherent panoramic geometry than MotionClone and exhibits fewer boundary artifacts than GoWithTheFlow.

\subsection{Ablation Study}
We conduct an ablation study on the full generated frames on the 360-1M test set and use a challenging various frame strides (1 to 3) to evaluate the impact on motion fidelity and overall video quality.
\cref{tbl:ablation} and \cref{fig:ablation} present the quantitative and qualitative results, respectively.

\subsubsection{Decoupled Motion Control.}
Our decoupled motion control module works as a plug-and-play component that can be easily integrated into existing motion-controllable video generation methods.
To demonstrate this, we incorporate it into the baseline GoWithTheFlow (GWTF) method, which improves most metrics, particularly those related to motion controllability.
We also ablate the module by removing from our full model, observing a significant drop across all metrics. This highlights its crucial role in enhancing motion fidelity.
In the very challenging case shown in \cref{fig:ablation}, where the target frame is temporally distant from the image condition and involves highly dynamic motion, the decoupled motion control improves the semantic consistency, e.g., generating a tree rather than a building in the center, by stabilizing the motion condition.
Interestingly, we find that the decoupled motion control also enhances loop-end continuity.
This is because it rotates the seam in the generated derotated video away from the boundary, presumably hacking the metric.
Therefore, in the following, we evaluate different loop consistency setups based on the ablated version without decoupled motion control.

\subsubsection{Loop Consistency.}
We further evaluate the proposed spherical noise warping and other techniques for improving loop consistency.
By ablating circular padding, the end continuity error increases dramatically, while other metrics remain similar.
Removing latent rotation and spherical noise warping leads to less increase in end continuity error, but also affects Flow EPE and PSNR, indicating that they also contribute to the overall video quality and motion control fidelity.
This is especially true for the spherical noise warping, as removing it leads to worse FVD and FID.
The qualitative results in \cref{fig:ablation} further confirm the results, where the circular padding mostly affects pixel level end continuity, which is hard to see in the generated video, while the removal of latent rotation and spherical noise warping leads to more severe artifacts in the seams.

\subsection{Applications Driven by Motion Control}
By combining decoupled motion control with spherical noise warping technique, our method can be easily applied to motion transfer and video editing tasks that require more robustness when motion control does not perfectly match the input image.
We achieve this by following a random degradation strategy~\cite{burgert2025go}, which blends a random amount of noise into the warped spherical noise when training.
During inference, we relax the motion condition by blending it with noise of equal strength.

\subsubsection{Motion Transfer.}
\cref{fig:transfer} presents the motion transfer results, where we transfer the motion from a source video to a drastically different target scene.
It is shown that our method can preserve the target scene's geometry and details while accurately following the motion of the source video.

\subsubsection{Video Editing.}
\cref{fig:editing} shows examples of video editing.
We use ChatGPT 4o's image generation to edit the first frame and the input prompt, and then generate the panorama videos based on the source motion.
The results show that our method can adapt existing motion to different geometries while preserving the motion control fidelity.

\section{Conclusion}\label{sec:conclusion}

In this paper, we presented \method, a novel framework for generating temporally coherent 360° panoramic videos with precise motion control from a single image.
We introduced a decoupled spherical motion-flow mechanism that enables more stable and high-fidelity motion control.
The spherical noise warping is introduced alongside other loop consistency techniques for seamless motion generation across panorama boundaries.
We also curated a large-scale dynamic panorama dataset with frame-level flow and pose annotations to support training and evaluation.
Extensive experiments across multiple datasets show that \method significantly outperforms prior methods in motion magnitude and fidelity, temporal coherence, and visual quality.

\noindent\textbf{Limitations.}
While the SLAM-based rotational estimation in our pipeline is robust under large motions, it may degrade when parallax is small---a limitation that could be mitigated by a learned derotation module~\cite{kim2021self}, which produces robust rotation outputs, reduces the need for reference videos, and presents a promising direction for future work.

\section*{Acknowledgments}
This research is supported by Building 4.0 CRC.

\bibliography{aaai2026}

\clearpage
\begin{center}
{\LARGE \textbf{Supplementary Material}}
\end{center}

\section{Implementation Details}

Our base model architecture is CogVideoX-5B-I2V~\cite{yang2025cogvideox}, an image-conditioned, transformer-based video diffusion model.
It generates 49 frames at a resolution of $480 \times 720$.
To adapt it for panoramic video generation, we fine-tune the model using LoRA~\cite{hu2022lora} modules with rank 128, enabling efficient training and largely preserving the base model's prior knowledge.
During training, we inject spherical warped noise derived from the optical flow of source panoramic videos, allowing the model to learn motion-conditioned denoising.
To increase motion diversity, we apply random frame sampling strides between 1 and 3 during training.
As many panoramic videos have a vanishing point near the center, it introduces a positional bias and degrades performance when this is not the case in latent rotation steps at inference time.
To address this, we apply random rotations to the source videos during training and reduce the model's center bias.
The model is trained on our curated dataset using a batch size of 32 for 15k steps, optimized with AdamW~\cite{loshchilovdecoupled} at a learning rate of $2 \times 10^{-5}$.
We use 8 NVIDIA A100 GPUs for training, with each GPU processing a batch size of 4. The full training procedure takes approximately 2 days to complete.
At inference time, we set the rotation angle $\theta$ to $40^\circ$ and use DDIM~\cite{song2020denoising} with 30 diffusion steps. The value of $\theta$ is chosen so that the latent map width $W = 90$ is divisible by $360 / \theta$, ensuring that the rolling distance per rotation step is an integer.
After generation, the output resolution is resized to $480\times960$ to restore the panoramic aspect ratio.
We hold out 100 clips from the curated dataset as a test set.
To further evaluate the out-of-domain generalizability, we additionally select 100 random clips from the WEB360 dataset~\cite{wang2024360dvd} for evaluation.

\section{Detailed Dataset Curation Process}

\begin{figure*}[t]
    \centering
    \includegraphics[width=1.\linewidth, trim={0 0 0 0}, clip]{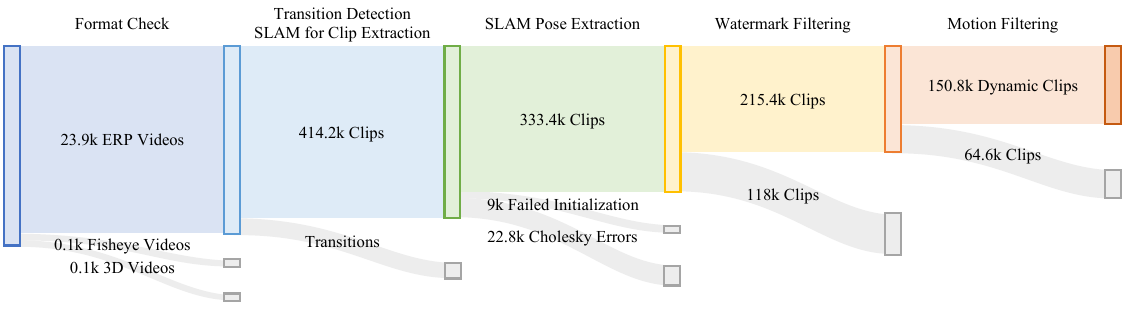}
    \caption{
        \textbf{Our proposed data curation pipeline.}
        We design a data curation pipeline to extract and clean dynamic clips from the 360-1M dataset.
        This results in a motion-rich panoramic video dataset, with around 150k clips annotated with camera poses.
    }\label{fig:dataset}
\end{figure*}

As described in the main paper, we construct a motion-rich panoramic video dataset by curating videos from 360-1M~\cite{wallingford2024image}, which contains a download list of 24.1k YouTube panoramic videos.
Our goal is to enable motion-controllable panorama video generation via motion decoupling.
To this end, as shown in~\cref{fig:dataset}, we design a robust preprocessing pipeline to extract and clean dynamic clips by format checking, removing transitions, filtering watermarks, estimating camera poses, and discarding low-motion segments.

\subsubsection{Format Check.}

The raw 360-1M dataset contains various non-equirectangular projection (ERP) formats, such as stereo 3D (left-right eye) videos and fisheye recordings.
For stereo 3D videos, we observe that they contain two nearly identical halves side-by-side. We compute the average SSIM~\cite{ssim} between the left and right halves as a 3D similarity score.
For fisheye videos, they typically have circular imaging regions surrounded by black borders.
We apply a circular mask to retrieve the out-of-bound areas and calculate a percentage of black pixels as fisheye score, considering gray-scale pixel values below 0.02 as black.
Videos with either a 3D score or fisheye score above 0.7 are discarded.
This step reduces the dataset from 24.1k to 23.9k valid ERP videos.

\subsubsection{Transition Detection.}
As training video generation models requires continuous video clips, we use a python library \texttt{scenedetect}~\cite{scenedetect2024} to detect scene transitions in the videos.
This library uses a threshold-based method to detect abrupt changes in the video frames.

\subsubsection{SLAM for Clip and Pose Extraction.}

While the \texttt{scenedetect} library can detect abrupt transitions, it often fails to handle soft transitions such as fade-ins and fade-outs. To address this limitation and obtain reliable scene boundaries, we leverage the tracking module of OpenVSLAM~\cite{sumikura2019openvslam}, a visual SLAM system tailored for panoramic videos. The SLAM system estimates camera trajectories based on keypoint matching; when tracking is lost, it typically indicates a significant change in the scene content. We treat such tracking failures as implicit scene transitions and remove frames from 1 second before to 1 second after each detected transition to ensure temporal smoothness and eliminate any residual blending artifacts.
Using these criteria, we segment the videos into 414.2k continuous clips, each ranging from 3 to 10 seconds in duration. We then run OpenVSLAM again on each resulting clip to extract the corresponding camera poses. During this process, 9k clips fail to initialize tracking due to insufficient camera motion, and 22.8k additional clips fail during pose graph optimization because of Cholesky decomposition errors. After filtering out these failure cases, we obtain a total of 333.4k clips with valid and high-quality camera pose annotations.

\subsubsection{Watermark Filtering.}
One key limitation of the WEB360 dataset is that many of its videos are sourced from a YouTube channel with persistent watermarks embedded in the 360° footage. This introduces a significant bias in the training data, often leading the model to hallucinate similar watermark patterns in generated frames. We observe that a subset of videos in 360-1M exhibits the same issue. To mitigate this, we employ a pretrained watermark detection model~\cite{watermark}, which assigns a score between 0 and 1 to each frame. For each video clip, we sample five frames, compute their individual scores, and take the average as the clip-level watermark score. Clips with an average score above 0.3 are discarded, resulting in a cleaned video set of 215.4k clips.

\subsubsection{Motion Filtering.}
To ensure the dataset is composed of motion-rich clips, we compute optical flow using the Farneback algorithm~\cite{farneback2003two}, implemented via OpenCV~\cite{opencv_library}. To reduce computational cost, each video frame is resized such that its minimum dimension is 256 pixels, and we process a subset of the dataset by sampling every 10th clip. For each clip, we calculate the average optical flow magnitude across all frames, which serves as a motion score. Clips with motion scores below a threshold of 2.0 are discarded to remove static or low-motion content.
After motion filtering, we retain a total of 150.8k dynamic clips.

\subsubsection{}
For training our motion decoupling method, we further use camera pose filter to encourage dynamic camera movements, and random select maximum of 5 clips from each video to balance semantic diversity.
We then generate text captions using the video captioning model CogVLM~\cite{wang2024cogvlm}.
Since several stages of the pipeline, especially the SLAM-based pose estimation, are computationally intensive, we implement the entire processing workflow in a distributed fashion on a large-scale CPU cluster.
The full curation process takes approximately 3 days to complete using around 1,000 CPU cores.
The final dataset comprises approximately 150k high-quality panoramic video clips, each annotated with camera poses.
We plan to release both the dataset and the full curation pipeline to support future research in panoramic video generation, 3D reconstruction, and related domains.


\end{document}